\documentclass[conference]{IEEEtran}
\IEEEoverridecommandlockouts

\usepackage{fancyhdr}
\fancypagestyle{withfooter}{

\fancyfoot[C]{\footnotesize Accepted to the Novel Approaches for Precision Agriculture and Forestry with Autonomous Robots IEEE ICRA Workshop - 2025}
}
\usepackage{etex}
\usepackage[latin1]{inputenc} 

\usepackage{cite}
\usepackage{amsmath,amssymb,amsfonts}
\usepackage{algorithmic}
\usepackage{graphicx}
\usepackage{textcomp}
\usepackage[dvipsnames]{xcolor}
\def\BibTeX{{\rm B\kern-.05em{\sc i\kern-.025em b}\kern-.08em
    T\kern-.1667em\lower.7ex\hbox{E}\kern-.125emX}}

\usepackage{hyperref}
\usepackage{subfig}

\usepackage{bm} 
\def\r#1{\bm{#1}} 
\def\la#1{\label{eq:#1}}

\usepackage{cancel}
\usepackage{float}

\begin{document}

\title{Robust 2D lidar-based SLAM in arboreal environments without IMU/GNSS*
\thanks{*This project has been supported by the  National Agency of Research and Development (ANID) under grants Fondecyt 1220140, ANID AFB240002 Basal Project and Doctoral Grant 21212303.}
}

\author{%
\IEEEauthorblockN{Paola Nazate-Burgos}
\IEEEauthorblockA{\textit{Department of Electrical Engineering} \\
\textit{Pontificia Universidad Católica de Chile}\\
Santiago, Chile \\
pjnazate@uc.cl}
\and
\IEEEauthorblockN{Miguel Torres-Torriti}
\IEEEauthorblockA{\textit{Department of Electrical Engineering} \\
\textit{Pontificia Universidad Católica de Chile}\\
Santiago, Chile \\
dammr@uc.cl}
\and
\IEEEauthorblockN{Sergio Aguilera-Marinovic}
\IEEEauthorblockA{\textit{Department of Mechanical  Engineering} \\
\textit{Pontificia Universidad Católica de Chile}\\
Santiago, Chile \\
sfaguile@uc.cl}
\and
\IEEEauthorblockN{Tito Arévalo}
\IEEEauthorblockA{\textit{Department of Mechanical  Engineering} \\
\textit{Pontificia Universidad Católica de Chile}\\
Santiago, Chile \\
tito.arevalo@uc.cl}
\and
\IEEEauthorblockN{Shoudong Huang}
\IEEEauthorblockA{\textit{School of Mechanical and Mechatronic Engineering} \\
\textit{University of Technology Sydney}\\
New South Wales, Australia \\
shoudong.huang@uts.edu.au}
\and
\IEEEauthorblockN{Fernando Auat Cheein}
\IEEEauthorblockA{\textit{Department of Engineering} \\
\textit{Harper Adams University}\\
Newport, England, UK \\
fauat@harper-adams.ac.uk}
} 

\maketitle

\begin{abstract}
Simultaneous localization and mapping (SLAM) approaches for mobile robots remains challenging in forest or arboreal fruit farming environments, where tree canopies obstruct Global Navigation Satellite Systems (GNSS) signals. Unlike indoor settings, these agricultural environments possess additional challenges due to outdoor variables such as foliage motion and illumination variability. This paper proposes a solution based on 2D lidar measurements, which requires less processing and storage, and is more cost-effective, than approaches that employ 3D lidars. Utilizing the modified Hausdorff distance (MHD) metric, the method can solve the scan matching robustly and with high accuracy without needing sophisticated feature extraction. The method's robustness was validated using public datasets and considering various metrics, facilitating meaningful comparisons for future research.  Comparative evaluations against state-of-the-art algorithms, particularly A-LOAM, show that the proposed approach achieves lower positional and angular errors while maintaining higher accuracy and resilience in GNSS-denied settings. This work contributes to the advancement of precision agriculture by enabling reliable and autonomous navigation in challenging outdoor environments.

\end{abstract}


\section{Introduction}
Simultaneous Localization and Mapping (SLAM) has emerged as a key area of research in agricultural robotics, driven by its potential to enable autonomous navigation and environment reconstruction in precision farming applications. The effectiveness and applicability of SLAM techniques in this domain have been increasingly recognized in recent literature~\cite{Aguiar2022, Debeunne2020, Liu2022}. While semantic and topological SLAM methods have demonstrated robustness in structured indoor environments, recent studies have begun exploring their extension to more complex outdoor settings, such as forestry environments~\cite{Liu2022}.



\begin{figure}[H]
	\centering
	\includegraphics[width=\columnwidth]{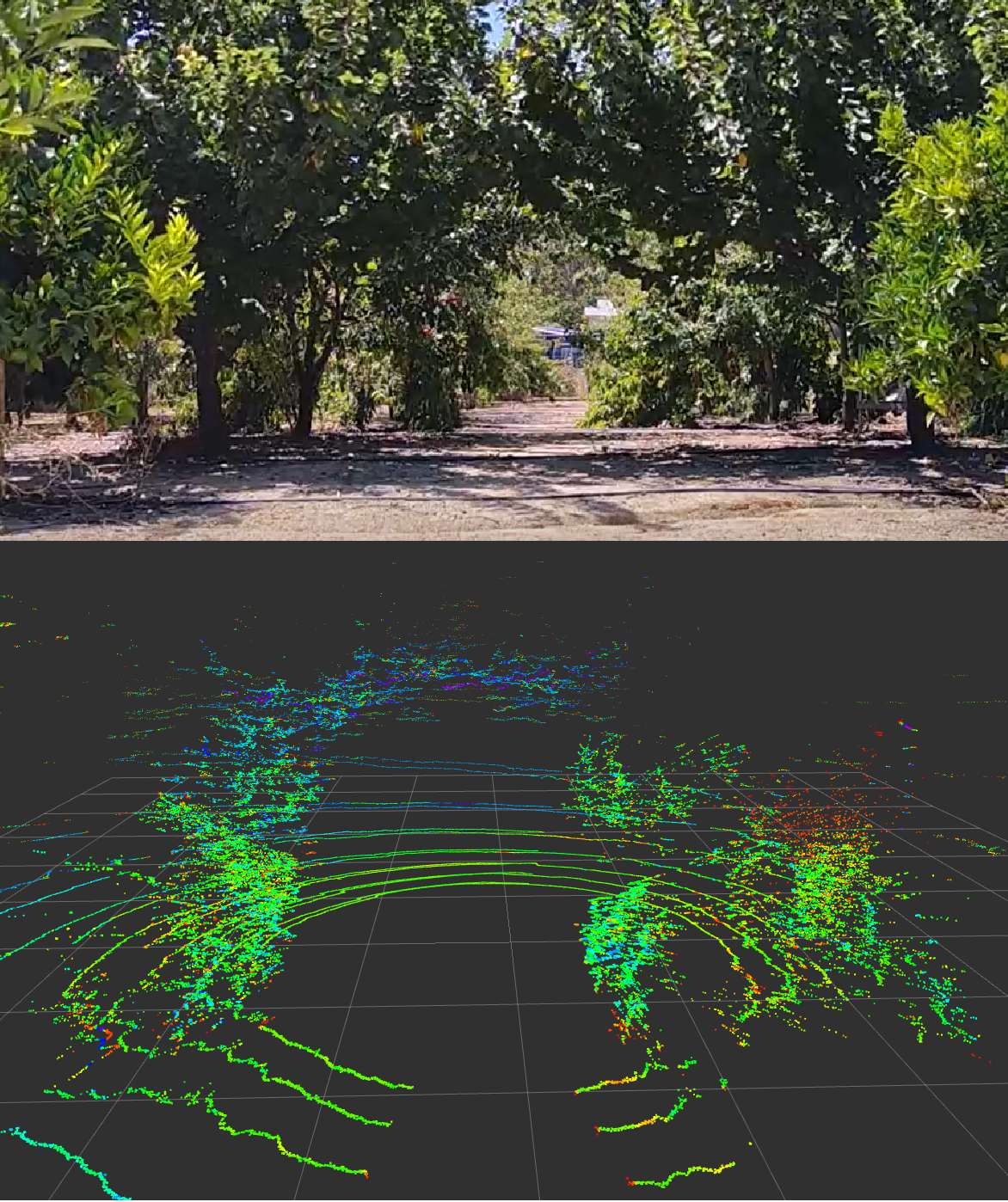}
	\caption{Fruit orchard with dense canopies: RGB camera view (top) and 3D radar scan (bottom).}
	\label{fig:intro}
\end{figure}

This paper introduces a SLAM methodology tailored specifically for arboreal and orchard environments, as illustrated in Fig.\ref{fig:intro}, where dense tree canopies frequently obstruct Global Navigation Satellite System (GNSS) signals, rendering high-precision, centimeter-level positioning infeasible. A fundamental challenge in these environments is the difficulty in identifying distinctive and repeatable features for reliable data association, due to the homogeneity and repetitive geometry of tree trunks and branches\cite{Cremona2022, Debeunne2020}. One of the earliest implementations of SLAM in tree groves was presented in~\cite{AuatCheein2011}, employing landmark-based SLAM with an Extended Information Filter (EIF-SLAM). However, the effectiveness of such approaches is often constrained by the ambiguity in landmark detection and association. To address these challenges, we propose a scan matching-based SLAM framework that robustly aligns 2D lidar scans without requiring an explicit data association phase. The method is designed to be resilient to outliers, commonly introduced by foliage and branching structures, and builds upon earlier work by the authors on robust localization techniques in highly repetitive and geometrically ambiguous environments, such as mining tunnels~\cite{TorresTorriti2022}. For a broader overview of SLAM applications in agricultural robotics, readers are referred to~\cite{TorresTorriti2022a}.


The principal contributions of this work can be summarized as follows: (i) A lidar-based SLAM framework tailored for arboreal environments, specifically designed to operate without inertial correction, enabling robust localization and mapping in forestry and orchard conditions without relying on IMU data; (ii) A novel localization approach based on Modified Hausdorff Distance (MHD) for scan matching, which effectively captures the spatial distribution of tree trunks and foliage, facilitating accurate alignment without the need for explicit feature extraction or data association; (iii) Full applicability in GNSS-denied and IMU-free settings, demonstrating resilience in environments where satellite navigation and inertial sensing are unreliable or unavailable due to canopy occlusion or cost constraints; (iv) Comprehensive experimental validation using multiple real-world agricultural datasets (CitrusFarm, Bacchus, and Pullally), as well as a controlled field environment, with performance evaluated across a wide range of positional, angular, local consistency, and map accuracy metrics; and (v) Deployment on a quadruped robotic platform (Unitree Go1), illustrating the method's robustness and feasibility in navigating uneven and unstructured terrains typical of agricultural and forestry applications.

This paper is organized as follows. 
Section 2 presents the proposed approach for SLAM and tree scan matching using a the modified Hausdorff distance (MHD) metric. The third section summarizes and discusses the 
main results obtained in controlled and real-world environments with a Unitree Go1 quadrupped robot and a Velodyne VLP-16. Finally, the conclusions are mentioned in Section 4.




\section{Proposed Method}

\begin{figure}
	\centering
	\includegraphics[width=\columnwidth]{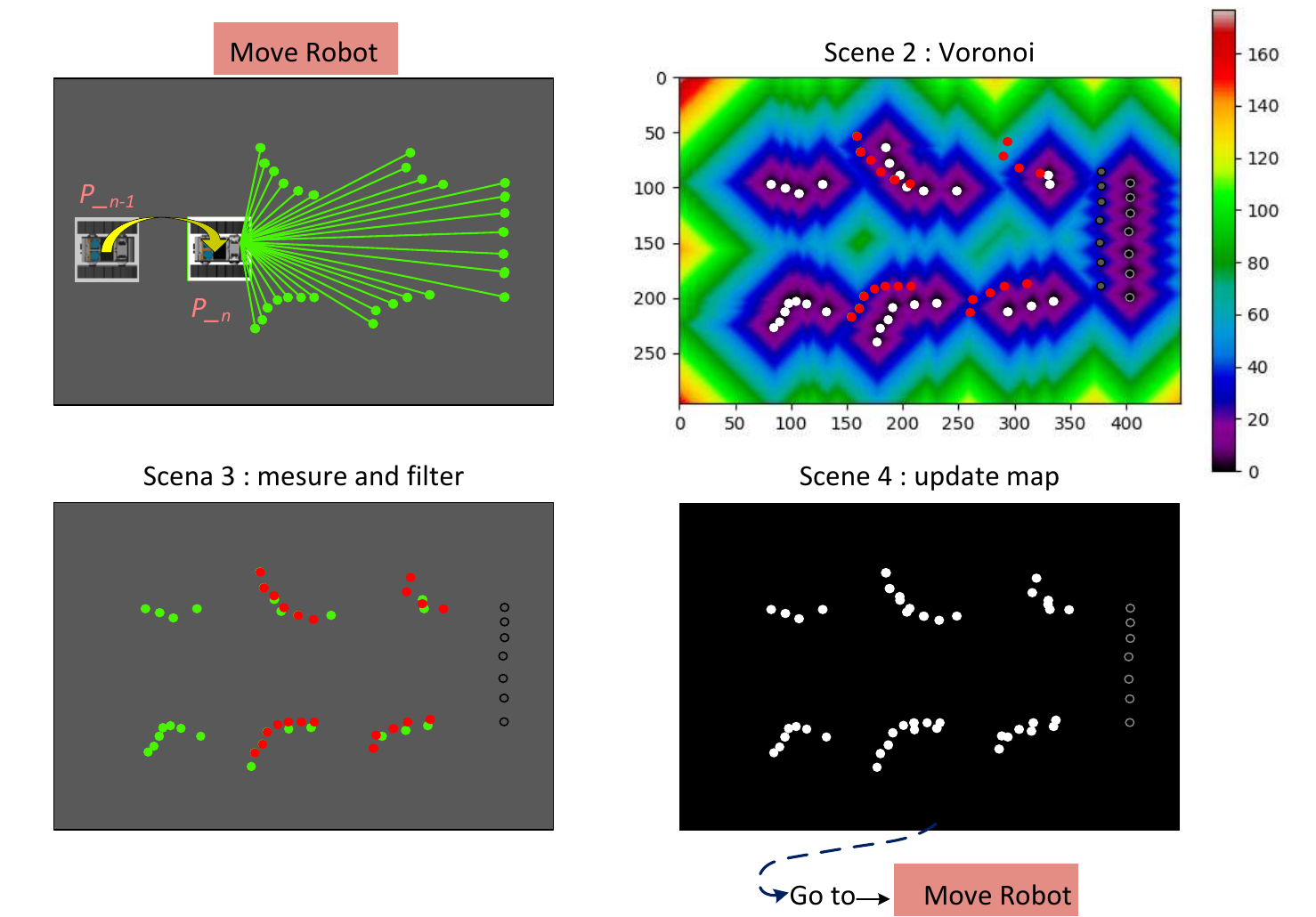}
	\caption{Data filtering and scan matching with MHD.}
	\label{fig:metodology}
\end{figure}
%
	%
		%
		%
			%

The proposed SLAM framework aims to estimate the pose $\r{q}_t=[x_t, y_t, \theta_t]$ of a mobile platform and concurrently construct or update a map $\mathcal{M}$ of the surrounding environment. The methodology is structured into three primary components: data processing, scan matching, and state/map updating.

\begin{figure}[h!]
	\begin{center}
		\includegraphics[width=\columnwidth]{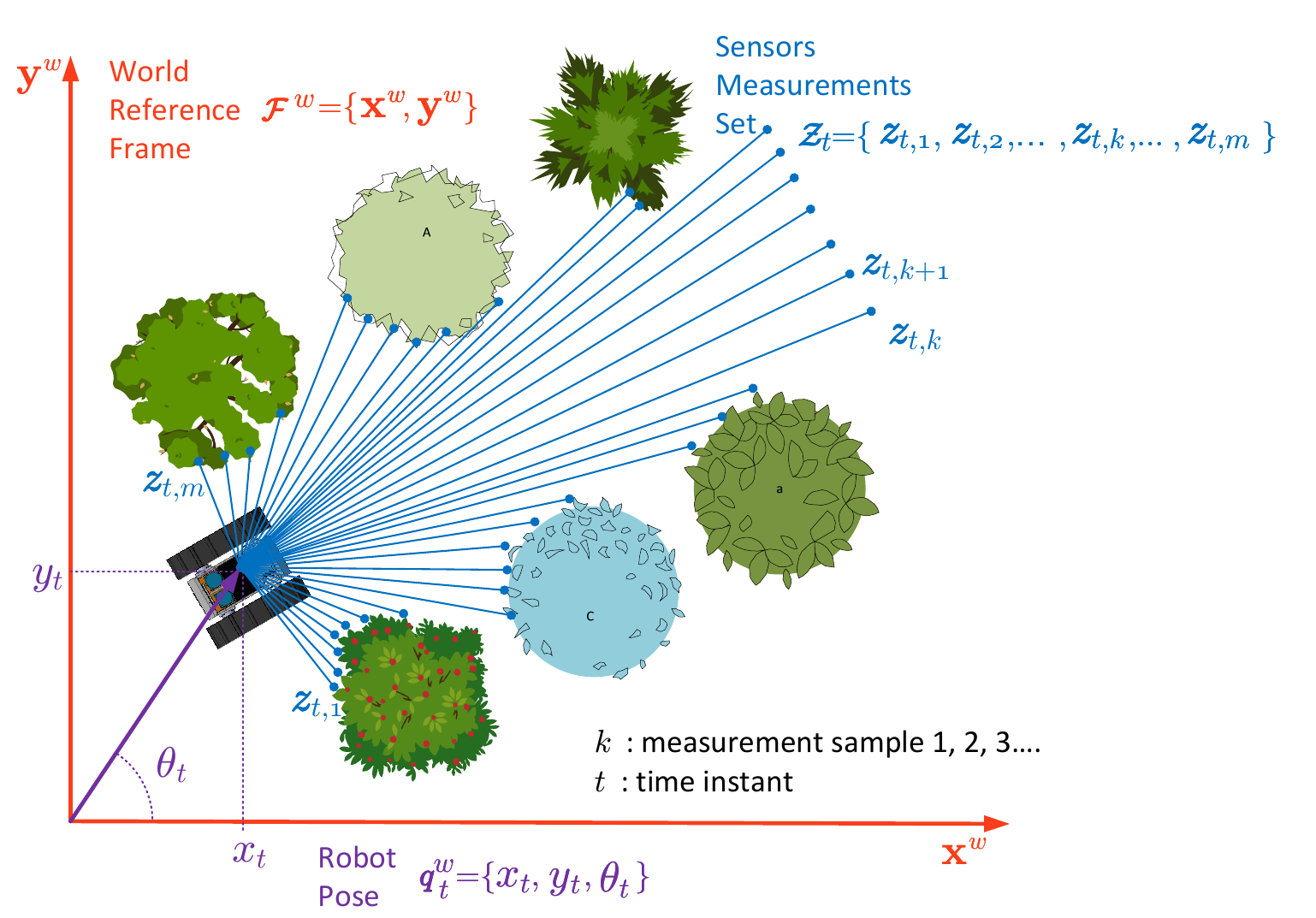}
	\end{center}
	\caption{The mobile robot's pose $\r{q}_t$ and lidar measurements $\r{z}_{t,k}$ in an agricultural environment.} \label{fig:notation}
\end{figure}

\subsection{Localization and State Representation}
The pose $\r{q}_t$ represents the robot's position $(x_t, y_t)$ and orientation $\theta_t$ in a 2D Cartesian world reference frame $\mathcal{F}_0$ at time $t$, shown in Fig.~\ref{fig:notation}. The system's state is equivalently defined by the pose:
$$\r{x}_t\stackrel{def}{=}\r{q}_t$$

\subsection{Motion Model}
The robot is modeled as a differential-drive system with state evolution governed by the nonlinear kinematic model:

\begin{eqnarray*}
&
	\dot{\r{x}}_t = \r{f}(\r{x}_t,\r{u}_t)=
	\left [
	\begin{array}{c}
		\dot{x}_t\\
		\dot{y}_t\\
		\dot{\theta}_t\\
	\end{array}
	\right ]
	 = 
	\left [
		\begin{array}{c}
			v_t\cos(\theta_t)\\
			v_t\sin(\theta_t)\\
			\omega_t
		\end{array}
		\right ],\ \r{x}_0 = \r{x}_\text{init}
& \la{sys1}
\end{eqnarray*}
where $\r{u}_t = [v_t, \omega_t]$ comprises the linear and angular velocities, respectively.

\subsection{Map Representation}
The map $\mathcal{M}$ is modeled as a probabilistic occupancy grid:
$$\mathcal{M}\stackrel{def}{=}\left \{\r{m}_{i,j},i= 1,2, \ldots N_i, j = 1,2,\ldots N_j\right \}.$$
Each cell $\r{m}_{i,j}$ encodes the occupancy likelihood, updated incrementally using lidar observations interpreted as conditional probabilities of occupancy given current sensor data. This formulation allows for consistent integration of noisy measurements over time.

\subsection{Lidar measurements processing}

\begin{figure*}[h!]
	\begin{center}
		\includegraphics[width=1.0\textwidth]{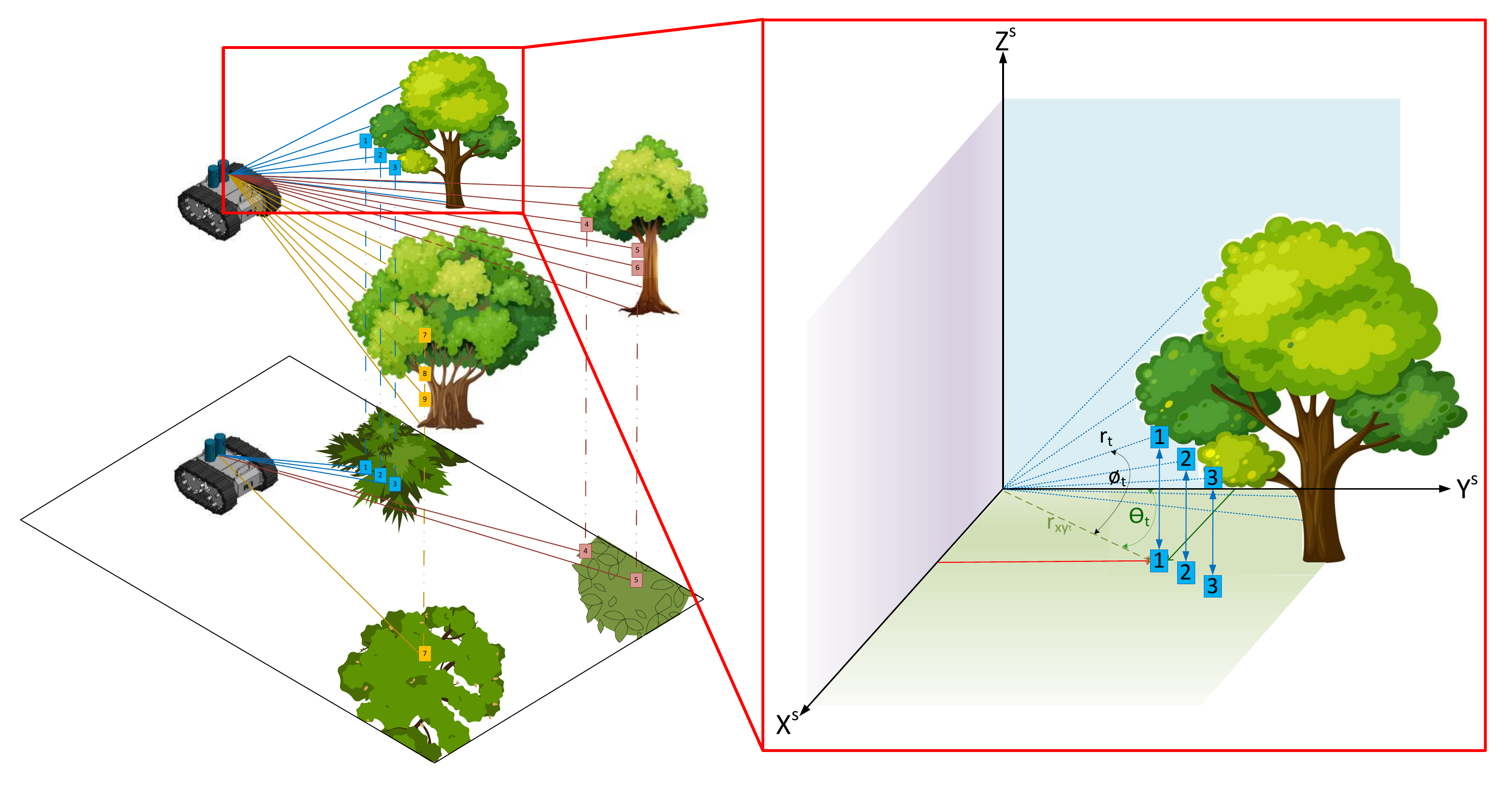}
	\end{center}
	\caption{Slicing of 3D lidar measurements and 2D projection onto ``ground'' plane.} \label{fig:flatten_measure}
\end{figure*}

Lidar data is initially represented as a 3D point cloud in the sensor's frame $\mathcal{F}_s$:
\begin{eqnarray*}
	\mathcal{Z}_t^s &=&\left \{\left (r_{t,k}^s,\theta^s_{t,k},\phi^s_{t,k}\right ),\ k=1,2,\ldots,N_s\right \}
\end{eqnarray*}
These are converted to Cartesian coordinates:
\begin{eqnarray*}
	x_t^s &=& r_{t,k}^s\cos(\theta_{t,k}^s)\cos(\phi_{t,k}^s),\\
	y_t^s &=& r_{t,k}^s\sin(\theta_{t,k}^s)\cos(\phi_{t,k}^s),\\
	z_t^s &=& r_{t,k}^s\sin(\phi_{t,k}^s),
\end{eqnarray*}
To reduce dimensionality and focus on relevant features, a horizontal slice is extracted (e.g., $z \in [0.0, 0.2]$ m). This 3D slice is projected into a 2D polar scan and discretized into angular bins, selecting the minimum range per bin as illustrated in Fig.~\ref{fig:flatten_measure}.

The resulting filtered scan is:
\begin{eqnarray*}
	\mathcal{Z}_{t_{filtered}}^s &=&\left \{\left (r_{{t}_{f},k}^s,\theta^s_{t,k} \right ),\ k=1,2,\ldots,N_s\right \}
\end{eqnarray*}
These are transformed into global coordinates using the current pose estimate $(x_t, y_t, \theta_t)$:
\begin{eqnarray*}
	x_t^w &=& x_t+r_{t_{f},k}^s\cos(\theta_{t,k}^s+\theta_t),\\
	y_t^w &=& y_t+r_{t_{f},k}^s\sin(\theta_{t,k}^s+\theta_t).
\end{eqnarray*}
where $R(\theta_t)$ is the 2D rotation matrix.

\subsection{Scan Matching via Modified Hausdorff Distance (MHD)}

 the distance transform of the map $\mathcal{M}$ is computed first. The lidar scan is superimposed on the distance transform at an initial guess of the location.  Rotation and translation operations are performed to find the rotation and translation that minimize the MHD.  This process yields a pose estimate vector:
\begin{equation}
	\hat{q}^* = (\hat{x}^*,\hat{y}^*,\hat{\theta}^*)\ =\ 
	\arg\min_{(\hat{x},\hat{y},\hat{\theta})} 
	\bar{h}_{k}(\mathcal{M},\mathcal{M}_t(\hat{q},\mathcal{Z}_{t}^{w})) \label{eq:scan_matching}
\end{equation}
where $\bar{h}_{k}$ is a distance measure with the modified Hausdorff distance computed with the $k$ best matching object coordinates between  the reference Map $\mathcal{M}$ and the observed elements of the environment in measurements set $\mathcal{M}_t(\hat{q},\mathcal{Z}_{t}^{w})$.

Initially the map is not known, and the first scan $\mathcal{Z}_0$ is treated as the initial map, i.e. $\mathcal{M}_0=\mathcal{Z}_0$.  As the robot moves, subsequent scans $\mathcal{Z}_t$ at time $t$ are matched to the map $\mathcal{M}_{t-1}$ and integrated into the updated map $\mathcal{M}_t$:
\begin{equation}
	\mathcal{M}_{t}(\hat{q}, \mathcal{Z}_{t}^{w})=\mathcal{M}_{t-1}\cup \left\{\hat{z}(\hat{q}, \mathcal{Z}_{t,k}^{w} ):k=1,2,\ldots,N_{s}\right\}
\end{equation}

\subsection{Pose Refinement via Extended Kalman Filter (EKF)}
To smooth the estimated trajectory, the pose from scan matching $\hat{\r{q}}_t$ is 
fused with the motion model using an EKF:
\begin{itemize}
\item Prediction step:
\begin{eqnarray}
	\r{x}_{t|t-1} &=& \r{f}(\hat{\r{x}}_{t-1},\r{u}_{t-1})\\
	\r{P}_{t+1|t} &=& \r{A}_t \r{P}_{t|t} \r{A}^T_t + \r{G}_t \r{Q}_t \r{G}^T_t, \r{P}_{0|0}\equiv \r{P}_0\\
	\r{S}_{t+1|t} &=& \r{C}_t \r{P}_{t+1|t} \r{C}^T_t +\r{H}_t \r{R}_t \r{H}^T_t \nonumber
\end{eqnarray}
\item Update step (using scan match result as measurement):
\begin{eqnarray}
	\r{e}_{t+1} &=& \r{z}_{t+1} -  \r{z}_{t+1|t} \\
	\r{K}_{t+1} &=& \r{P}_{t+1|t} \r{C}^T_t + \r{S}^{-1}_{t+1|t}\\
	\r{x}_{t+1|t+1} &=& \r{x}_{t+1|t} + \r{K}_{t+1} \r{e}_{t+1} \\
	\r{P}_{t+1|t+1} &=& \r{P}_{t+1|t} - \r{K}_{t+1} \r{S}_{t+1|t} \r{K}^T_{t+1}
\end{eqnarray}

\end{itemize}

\subsection{Map and State Update}
Following pose correction, the occupancy grid map is updated using a recursive Bayesian formulation:
\resizebox{\columnwidth}{!}{%
	\parbox{1.1\columnwidth}{%
		\begin{eqnarray}
			&&p(\r{x}_{t}|\mathcal{M}_{t},\mathcal{Z}_{0:t},\r{u}_{0:t-1})
			\ =\ \nonumber \\
			&&\ \eta
			p(\mathcal{Z}_{t}|\r{x}_{t},\mathcal{M}_{t},\r{u}_{0:t-1})
			\ \cdot \nonumber \\
			&& \sum_{\r{x}_{t-1}} 
			p(\r{x}_{t}|\r{x}_{t-1},\r{u}_{t-1})
				p(\r{x}_{t-1},\mathcal{M}_{t-1}|\mathcal{Z}_{0:t-1},\r{u}_{0:t-2})
		\end{eqnarray}
}}

\resizebox{\columnwidth}{!}{%
	\parbox{1.2\columnwidth}{%
		\begin{eqnarray}
			p(\mathcal{M}_{t}|\r{x}_{t},\mathcal{Z}_{0:t},\r{u}_{0:t-1})
			&=&
			\sum_{\mathcal{M}_{t-1}} 
			p(\mathcal{M}_{t}|\r{x}_{t},\mathcal{M}_{t-1},\mathcal{Z}_{0:t},\r{u}_{0:t-1})
			\cdot\nonumber\\
			&&
			\sum_{\r{x}_{t-1}}
			p(\r{x}_{t-1},\mathcal{M}_{t-1}|\mathcal{Z}_{0:t-1},\r{u}_{0:t-2})
		\end{eqnarray}
}}
The state-map joint posterior is then factorized as:
\resizebox{\columnwidth}{!}{%
	\parbox{1.1\columnwidth}{%
		\begin{eqnarray}
			p(\r{x}_{t},\mathcal{M}_{t}|\mathcal{Z}_{0:t},\r{u}_{0:t-1}) &=&
			p(\r{x}_{t}|\mathcal{M}_{t},\mathcal{Z}_{0:t},\r{u}_{0:t-1})\cdot\nonumber\\
			&&
			p(\mathcal{M}_{t}|\r{x}_{t},\mathcal{Z}_{0:t},\r{u}_{0:t-1}). \label{eq:joint_state_map_belief}
		\end{eqnarray}
}}

\begin{figure}[h!]
	\begin{center}
		\includegraphics[width=\columnwidth]{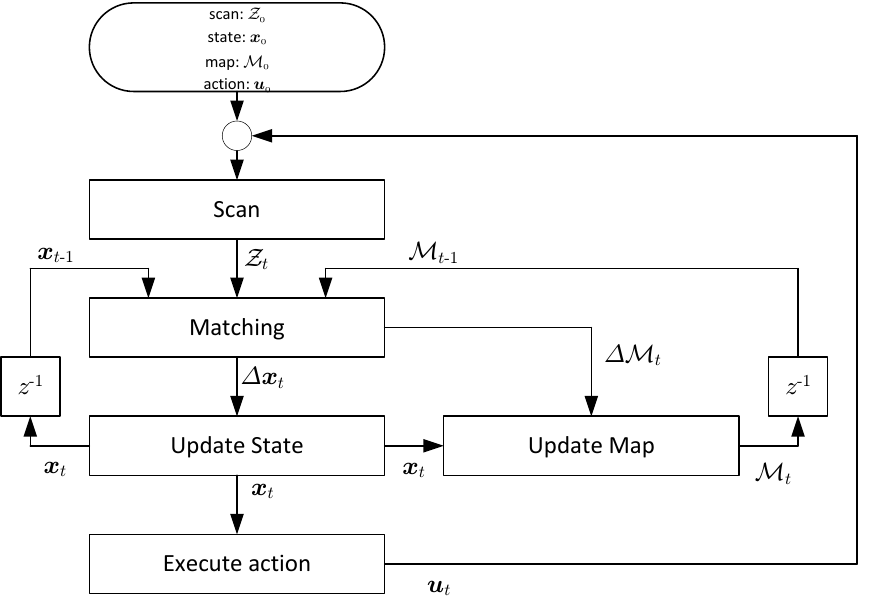}
	\end{center}
	\caption{Online SLAM navigation based on scan matching.} \label{fig:slam_matching}
\end{figure}

The flow diagram of Fig.~\ref{fig:slam_matching} summarizes the main steps of the proposed SLAM approach.

\section{Summary of Results}

\subsection{Controlled Field Environment}
\begin{figure}[ht]
	\centering
	\includegraphics[width=\columnwidth]{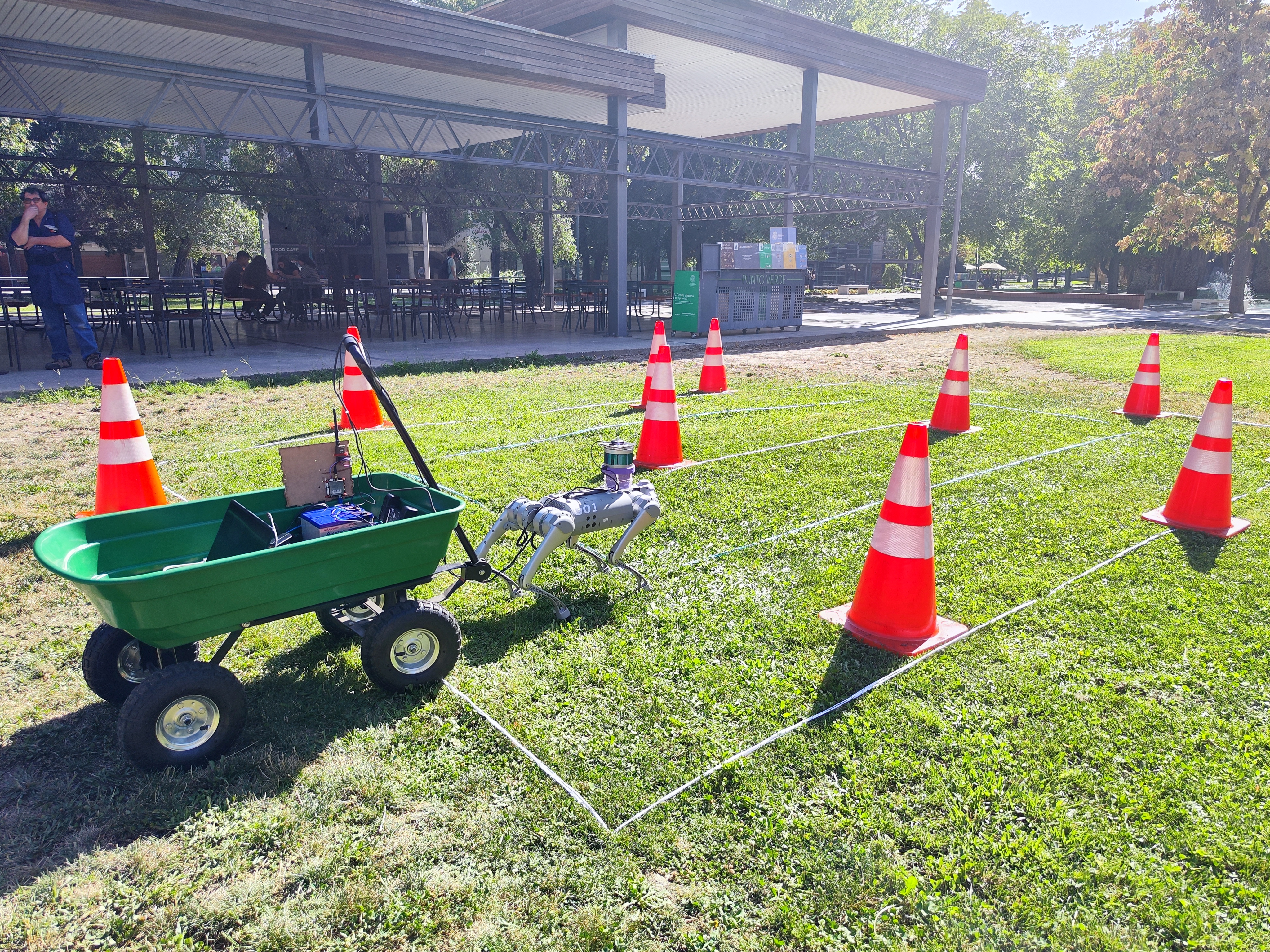}
	\caption{\centering Robot and traffic cones for control experiment and GNSS ground truth position acquisition.}
	\label{fig:fig_cones}
\end{figure}

To simulate real-world fruit forest conditions in a controlled and repeatable setting, experiments were conducted using a quadruped robot (Unitree Go1) navigating a 16~m closed-loop trajectory marked by traffic cones, simulating tree trunks; see Fig.~\ref{fig:fig_cones}. The terrain was grassy and uneven, and ground-truth pose was captured using centimeter-accurate GNSS-RTK (ArduSimple RTK3B).

\underline{Experiment Setup}
\begin{itemize}
\item Sensors: Velodyne VLP-16, ArduSimple RTK3B GNSS
\item Platform: Unitree Go1 quadruped
\item Trajectory: 6~m$\times$2~m rectangular path, repeated five times
\item Lighting: Daylight, minimal reflectance for optimal Lidar readings
\end{itemize}

\underline{Performance Evaluation}\\
The proposed method was benchmarked against A-LOAM (Advanced LOAM), using only lidar data (i.e., no IMU), across both pose and map accuracy metrics. The results are summarized in Table~\ref{tab:pes_cfe} and Table~\ref{tab:mes_cfe}. The proposed method consistently outperformed A-LOAM across all major metrics, particularly in absolute pose estimation. Both methods showed similar performance in local consistency metrics, such as incremental positional change errors: $\bar{e}_{\Delta pos}$: 0.03~m (ours) vs. 0.04~m (A-LOAM) and ${e_{RMS}}_{\Delta pos}$: 0.06~m (ours) vs. 0.05~m (A-LOAM); see Table~\ref{tab:pes_cfe} and Table~\ref{tab:mes_cfe}.

\begin{table}
\centering
\caption{Pose estimation results in controlled field environment.}
\label{tab:pes_cfe}
{\small
\begin{tabular}{lcc}\hline
Metric & Proposed Method & A-LOAM\\\hline
Mean Positional Error (m) & 0.08 $\pm$ 0.05 & 0.99 $\pm$ 0.78\\
Mean Angular Error ($^\circ$) & 0.12 $\pm$ 0.14 & 0.83 $\pm$ 0.99\\
RMS Positional Error (m) & 0.10 $\pm$ 0.01 & 1.37 $\pm$ 0.12\\
RMS Angular Error ($^\circ$) & 0.21 $\pm$ 0.01 & 1.21 $\pm$ 0.63\\
Loop Closure Error (m) & 0.15 $\pm$ 0.01 & 2.79 $\pm$ 0.02\\
\hline
\end{tabular}
}
\end{table}

\begin{table}
\centering
\caption{Map accuracy results in controlled field environment.}
\label{tab:mes_cfe}
{\small
\begin{tabular}{lc}\hline
Metric & Value\\\hline
Mean Map Error ($\bar{e}_{prom}$) & 0.28 $\pm$ 0.05 [pix]\\
Accuracy & 1.00 $\pm$ 0.01\\
F1 Score & 0.86 $\pm$ 0.02\\
Precision (PPV) & 0.92 $\pm$ 0.03\\
Sensitivity (TPR) & 0.80 $\pm$ 0.03\\
Specificity (TNR) & 1.00 $\pm$ 0.02\\
\hline
\end{tabular}
}
\end{table}

\subsection{Evaluation in Real Fruit Orchards}

The proposed SLAM method was evaluated on three real-world orchard datasets-CitrusFarm, Bacchus, and Pullally-chosen for their uneven terrain, dense foliage, and sensor challenges. These conditions introduce occlusions, illumination variability, and mapping difficulties, making them ideal for benchmarking robust SLAM solutions.  

The proposed method shows consistent reliability and robust performance across all datasets-particularly in GNSS-denied and IMU-free conditions.  The results are summarized in Table~\ref{tab:real_pos_error_citrus_bacchus}. It is notably resilient to outliers, foliage occlusion, and repetitive geometric features, outperforming A-LOAM in Bacchus and Pullally on most metrics.  A-LOAM performs better in structured environments with dense feature availability (e.g., CitrusFarm) but degrades in less regular, more dynamic settings.

\begin{figure}[h]
	\centering
	\begin{minipage}{0.45\textwidth} 
		\centering
		\subfloat[]{\includegraphics[width=\columnwidth]{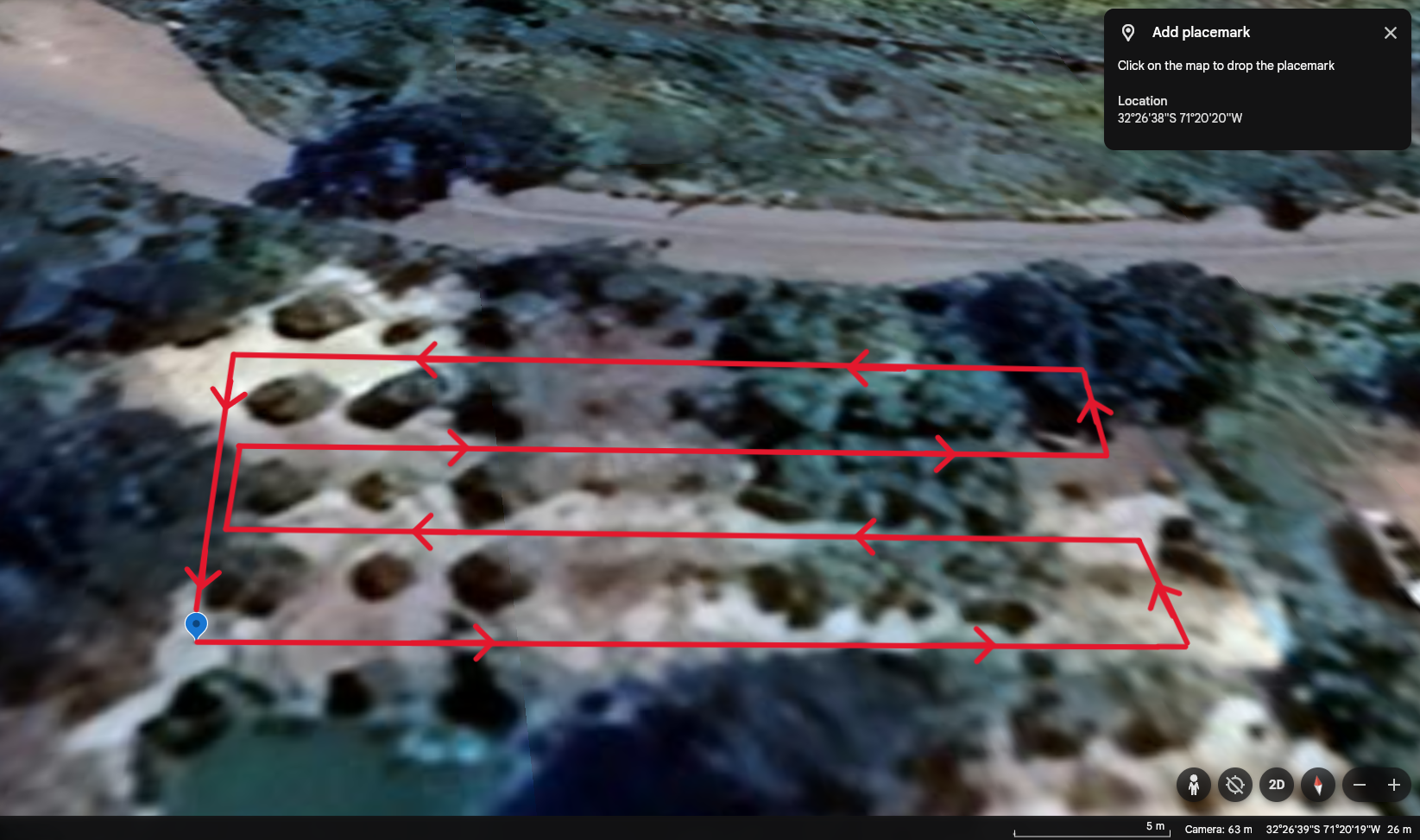} \label{fig:pullally_show_dataset_a}}
		\\
		\subfloat[]{\includegraphics[width=\columnwidth]{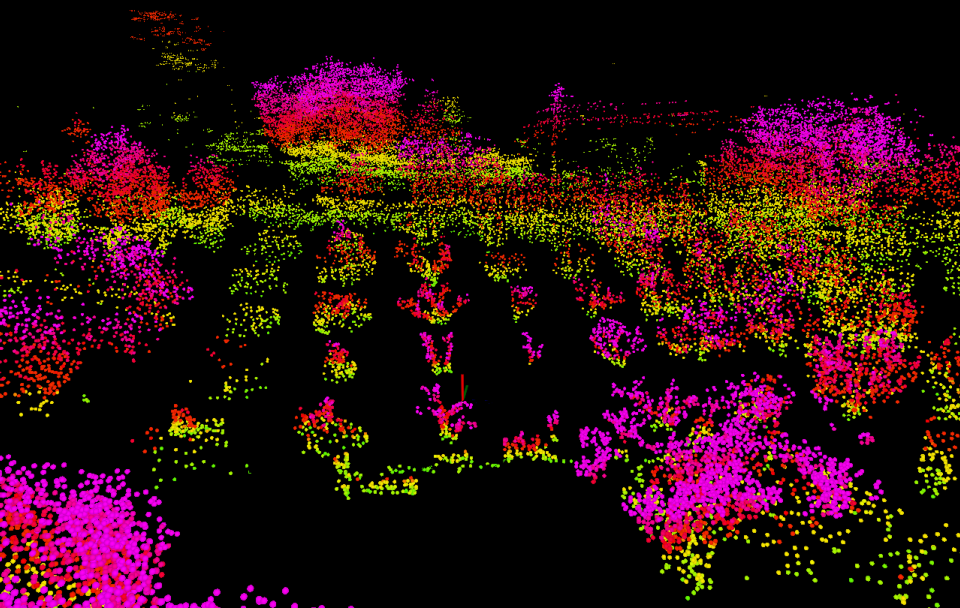} \label{fig:pullally_show_dataset_b}}
	\end{minipage}
	\caption{Pullally fruit orchard and the robot trajectory (red line) in (a),  and a lidar view of the fruit orchard (b).}
	\label{fig:pullally_show_dataset}
\end{figure}

\begin{figure}[h]
	\centering
	\begin{minipage}{1.\columnwidth} 
		\centering
		\subfloat[]{\includegraphics[width=\columnwidth]{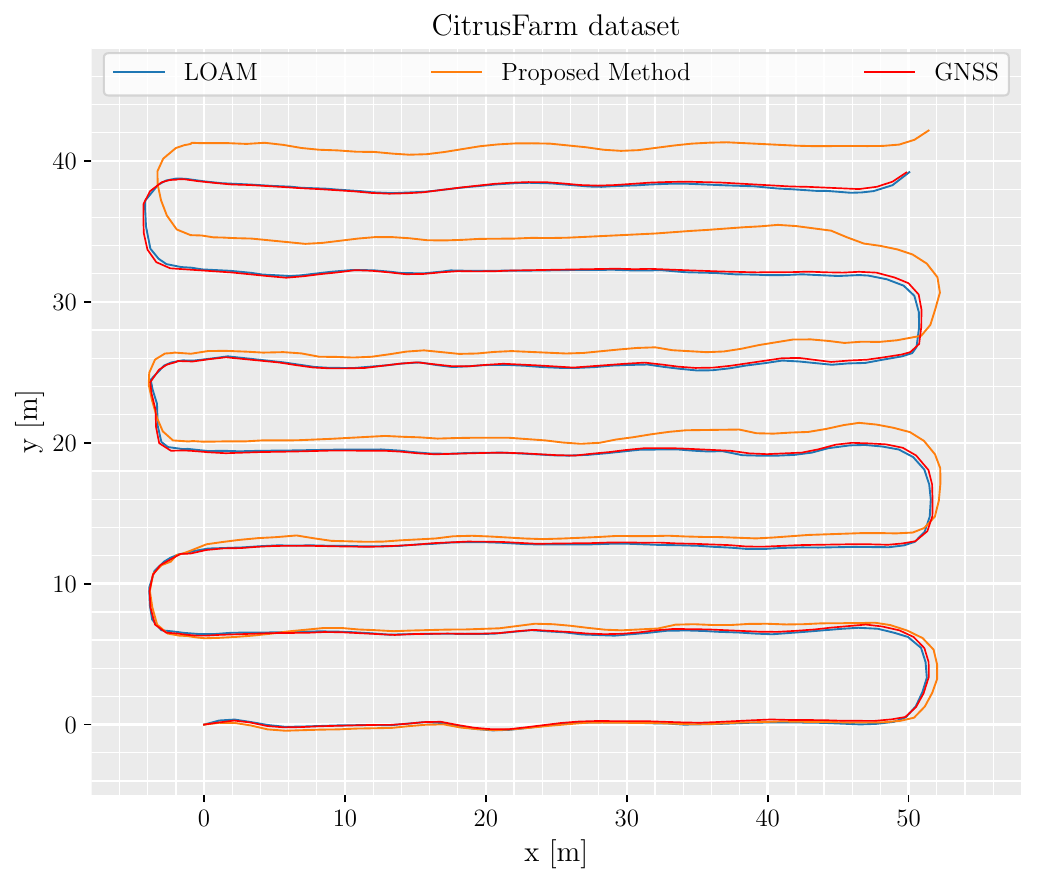} \label{fig:citrusfarm_results_comparison_a}}
	\end{minipage}
	\\
	\begin{minipage}{1.\columnwidth} 
		\centering
		\subfloat[]{\includegraphics[width=\columnwidth]{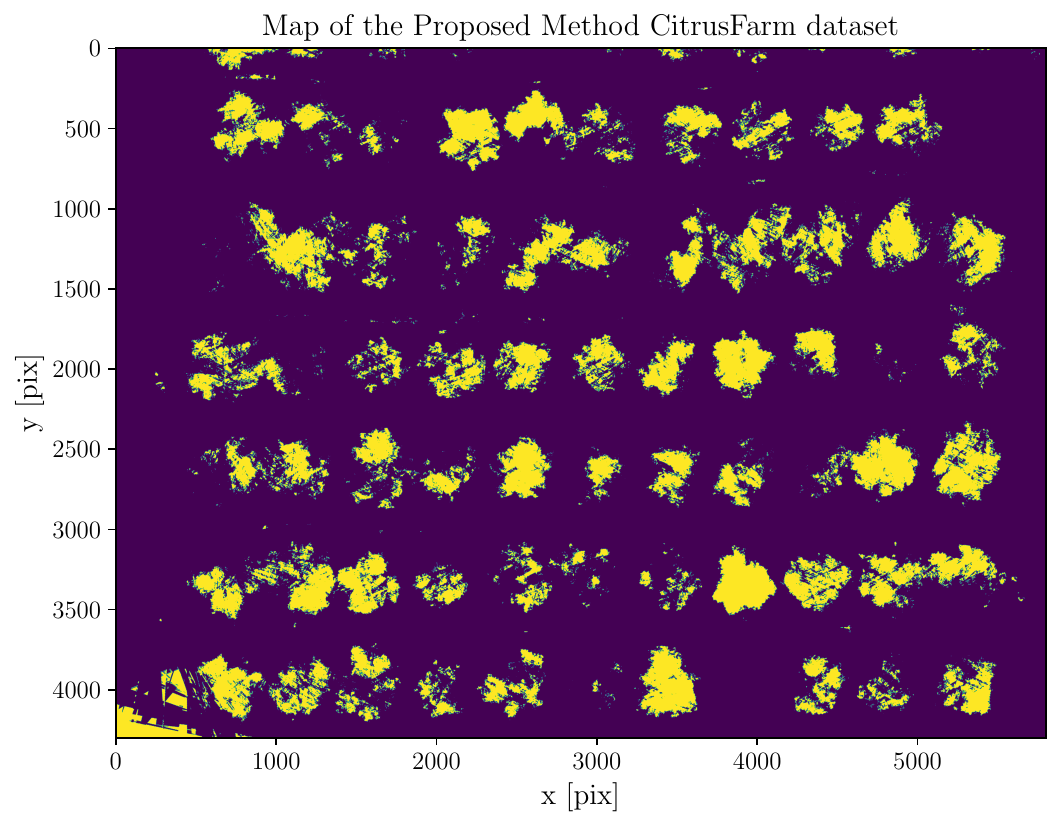} \label{fig:citrusfarm_results_comparison_b}}
	\end{minipage}
	\\
	\begin{minipage}{1.\columnwidth} 
		\centering
		\subfloat[]{\includegraphics[width=\columnwidth]{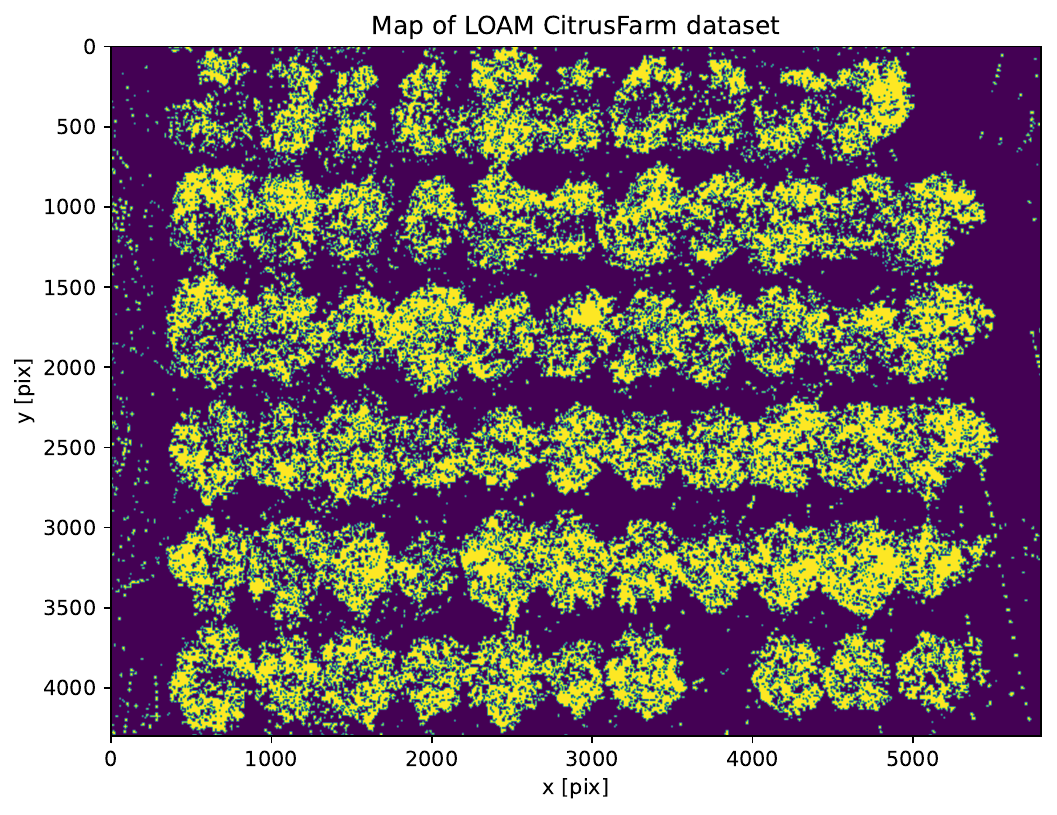} \label{fig:citrusfarm_results_comparison_c}}
	\end{minipage}
	\caption{CitrusFarm dataset results obtained with the proposed algorithm and A-LOAM comparing the estimated trajectory in fig. (a), and the map obtained with the proposed method with the A-LOAM algorithm. (b) and (c).}
	\label{fig:citrusfarm_results_comparison}
\end{figure}

\begin{enumerate}

\item CitrusFarm (UC Riverside; see Fig.~\ref{fig:citrusfarm_results_comparison}):
  \begin{itemize} 
	 \item Platform: Clearpath Jackal with Velodyne VLP-16 and SwiftNav RTK.
   \item Trajectory: 865.3~m across dense citrus clusters.
   \item Results: A-LOAM achieved lower positional errors (0.31~m vs. 1.48~m), but both methods showed comparable angular accuracy (0.06$^\circ$).
   \item Observation: A-LOAM outperformed due to its reliance on rich geometric features, which are abundant in this dataset.
  \end{itemize}
	
\item Bacchus (Vineyard, Greece):
  \begin{itemize}
	  \item Platform: 4WD4S Thorvald II with OUSTER OS1-16 and Trimble BX992 RTK.
    \item Trajectory: 102.3~m through trellis-configured grapevines.
    \item Results: The proposed method achieved better overall performance:
      \begin{itemize}
        \item Positional RMSE: 0.84~m (vs. 1.26~m for A-LOAM).
        \item Angular RMSE: 0.21$^\circ$ (vs. 2.48$^\circ$ for A-LOAM).
      \end{itemize}
		\item Observation: Trellis-like structures induced geometric degeneracy, degrading A-LOAM's performance. The proposed method remained robust to these structured occlusions.
  \end{itemize}
	
\item Pullally (Chile, dataset produced by authors; see Fig.~\ref{fig:pullally_show_dataset})\\
  \begin{itemize}
	  \item Platform: Unitree Go1 quadruped with Velodyne VLP-16 and Ardusimple RTK.
    \item Trajectory: 150~m through mixed-fruit plantations with seasonal foliage differences.
    \item Results: The proposed method vastly outperformed A-LOAM:
       \begin{itemize}
			   \item Mean positional error: 0.35~$\pm$ 0.13~m vs. 20.51~$\pm$~20.72~m.        
				 \item Angular RMSE: 0.18$^\circ$ vs. 3.84$^\circ$.
         \item Loop closure error: 0.25~m vs. 41.99~m.
         \item Observation: A-LOAM struggled significantly without lidar stabilization, while the proposed method maintained high accuracy despite the challenging terrain and platform dynamics.
				\end{itemize}
   \end{itemize}				
\end{enumerate}

\begin{table*}
	\caption{Pose errors of CitrusFarm and Bacchus datasets between the proposed method and ALOAM, where position ($pos$) and angular ($ang$) errors are respectively measured in $[m]$ and $[degrees]$.}
	\label{tab:real_pos_error_citrus_bacchus}
	\begin{center}
		\begin{tabular}{c|cc|cc|cc} 
			\hline
			Dataset & 
			\multicolumn{2}{c|}{CitrusFarm} & \multicolumn{2}{c|}{Bacchus}& \multicolumn{2}{c}{Pullally}\\[0.5ex] 
			\hline
			Method & Ours & A-LOAM & Ours & A-LOAM & Ours & A-LOAM\\
			\hline
			$\bar{e}_{pos}$ & 1.48 & 0.31 & 0.75 & 1.11 & 0.35 $\pm$ 0.13 & 20.51 $\pm$ 20.72 \\
			$\bar{e}_{ang}$ & 0.06 & 0.06 & 0.10 & 1.13 & 0.11 $\pm$ 0.12 & 3.09 $\pm$ 2.64 \\
			$e_{RMS_{pos}}$ & 1.83 & 0.36 & 0.84 & 1.26 & 0.37 $\pm$ 0.04 & 26.05 $\pm$ 4.18 \\
			$e_{RMS_{ang}}$ & 0.09 & 0.29 & 0.21 & 2.48 & 0.18 $\pm$ 0.02 & 3.84 $\pm$ 0.11 \\
			$\bar{e}_{\Delta pos}$ & 0.05 & 0.20 & 0.14 & 0.22 & 0.04 $\pm$ 0.04  & 0.05 $\pm$ 0.06  \\
			$\bar{e}_{\Delta ang}$ & 0.06 & 0.08 & 0.34 & 0.34 & 0.23 $\pm$ 0.21  & 1.81 $\pm$ 1.66 \\
			${e_{RMS}}_{\Delta pos}$ & 0.06 & 0.27 & 0.18 & 0.60 & 0.05 $\pm$ 0.01  & 0.07 $\pm$ 0.01 \\
			${e_{RMS}}_{\Delta ang}$ & 0.10 & 0.41 & 0.78 & 1.00 & 0.29 $\pm$ 0.03 & 2.32 $\pm$ 0.08 \\
			$\bar{e}_{\%}$ & 0.0026 & 0.0005 & 0.0073& 0.0105 & 0.001 $\pm$ 0.00 & 0.06 $\pm$ 0.01\\
			$loop_{e}$& 2.5605 & 0.1042 & 0.9591 & 1.1621 & 0.25 $\pm$ 0.00 & 41.99 $\pm$ 0.01 \\  
			\hline
		\end{tabular}
	\end{center}
\end{table*}

\section{Conclusions}

This work presents a robust, GNSS-independent SLAM framework specifically designed for challenging arboreal environments, such as fruit orchards and forests, where traditional localization approaches are hindered by occlusions, repetitive structures, and uneven terrain. The proposed method leverages a 2D lidar-based scan matching algorithm based on the Modified Hausdorff Distance (MHD), augmented with probabilistic state estimation via an Extended Kalman Filter (EKF) and a recursive Bayesian map update strategy.

The main contributions can be summarized in: (i) A novel lidar-based SLAM approach that operates effectively without GNSS or IMU data, using scan matching via MHD to avoid reliance on data association or feature extraction; (ii) Robust pose estimation that integrates a motion model and MHD-based measurements within an EKF framework, enabling consistent trajectory tracking in environments with limited observability; (iii) A complete and efficient mapping strategy that incrementally builds an occupancy grid from noisy lidar observations while preserving geometric consistency over time; (iv) Extensive validation across diverse conditions, including controlled environments, public datasets, and real-world field deployments on a quadruped robot, demonstrating the practical relevance and generalizability of the method.

On the other hand, the key experimental outcomes can be summarized in: (i) In controlled field experiments, the proposed method achieved sub-decimeter accuracy in both positional and angular estimates, outperforming A-LOAM in all global pose error metrics while maintaining local consistency; (ii) On public orchard datasets (CitrusFarm and Bacchus), the proposed method showed comparable or superior accuracy relative to A-LOAM, particularly in environments with structured degeneracy (e.g., trellised vineyards); (iii) In the challenging Pullally dataset, the method exhibited marked superiority, achieving robust localization despite foliage variability and sensor instability, where A-LOAM showed substantial performance degradation; (iv) The proposed method maintained high map accuracy and alignment across all tests, confirmed through statistical analysis of pixel-level map comparisons.

Overall, the proposed system demonstrates that accurate and reliable SLAM is achievable in complex outdoor agricultural environments without reliance on GNSS, IMU, or handcrafted features. These results support the use of the proposed framework in applications such as precision agriculture, autonomous orchard inspection, and long-term environmental monitoring.

Ongoing research is concerned with approaches for feature extraction from 3D lidar measurements in order to obtain complementary landmarks for localization, e.g. from canopy density and geometrical features. Another important research aspect is concerned with attenuating terrain disturbances on the measurements.  Vibrations transmitted to the sensors while the robot moves, as well as sudden changes in terrain slope increase the chances for matching errors. The effects of these disturbances are especially important when mounting the sensors on legged robots, in which intermittent foot-ground reaction forces occur while walking.  

\newpage

It is to be noted that the proposed approach was purposely designed to be minimalistic requiring only lidar measurements without IMUs. It is expected that the accuracy of the proposed may improve if IMU measurements are also included in the motion estimation.   
Hence, we also consider extending the current work to a more general setting,  accounting for full 6-DOF motion, and including additional comparisons with other recent popular SLAM approaches that depend on IMU measurements by design besides 3D lidar measurements.




\section*{Acknowledgement}

This project has been supported by the  National Agency of Research and Development (ANID) under Doctoral grant  21212303. Grant acknowledgements Fondecyt 1220140  and ANID AFB240002 Basal Project.

\bibliographystyle{IEEEtran}
\bibliography{IEEEabrv,slam}

\end{document}